\documentclass[letterpaper]{article} 
\usepackage{aaai2027}  
\usepackage[hyphens]{url}  
\usepackage{graphicx} 
\usepackage{natbib}  
\usepackage{caption} 
\usepackage{algorithm}
\usepackage{algorithmic}
\usepackage{amsmath}
\usepackage{amssymb}
\usepackage{booktabs}
\usepackage{newfloat}
\usepackage{listings}
\DeclareCaptionStyle{ruled}{labelfont=normalfont,labelsep=colon,strut=off} 
\floatstyle{ruled}
\newfloat{listing}{tb}{lst}{}
\floatname{listing}{Listing}

\usepackage{booktabs}

\title{QuantumMind: Constraint-Grounded Agentic Reasoning\\ for Speedup Analysis in Quantum Computing}

\author{
    Yijing Zuo\textsuperscript{\rm 1},
    Zhe Fu\textsuperscript{\rm 1},
    Zihan Nie\textsuperscript{\rm 2},
    Zhihui Zhu\textsuperscript{\rm 3},
    Haohan Wang\textsuperscript{\rm 1}
}

\affiliations{
    \textsuperscript{\rm 1}University of Illinois at Urbana-Champaign\\
    \textsuperscript{\rm 2}Rensselaer Polytechnic Institute\\
    \textsuperscript{\rm 3}Ohio State University
}

\nocopyright

\begin{document}

\maketitle

\begin{abstract}
Identifying a meaningful quantum speedup requires more than matching a
classical problem to a familiar quantum primitive: the claim must preserve the
task, respect access and output models, expose required promises, and remain
within a defensible complexity scope. We present QuantumMind, an auditable
agentic workflow for generating and conservatively screening
quantum-acceleration hypotheses. A fixed sequence of typed, role-specialized
actions formalizes the public task, analyzes structure and classical
bottlenecks, matches a source-linked registry of quantum primitives and
barriers, and constructs a scoped candidate scheme. A deterministic ten-check
validator assigns the authoritative verdict; completed states are then
compiled into a Quantum Acceleration Evidence Graph and passed through a
downward-only research screen that cannot strengthen the decision. We evaluate
QuantumMind against seven task-adapted prompting and agentic controls on 582
identical open-discovery tasks. Under the frozen Open-Discovery Score (ODS),
QuantumMind obtains 53.1 mean ODS, exceeding the strongest baseline by 17.3
points (48.2\% relative), and wins 355 of 582 paired tasks against that
baseline. It passes the graph audit on 99.8\% of tasks, compared with 43.6\%
for the strongest baseline, and ranks first in all seven task families. The
results indicate that typed state transitions and deterministic evidence
control contribute beyond fluent generation alone.
\end{abstract}

\section{Introduction}
\label{sec:introduction}

Quantum speedups arise when a computational problem satisfies the structural
and operational requirements of a quantum primitive. Factoring reduces to
period finding, unstructured witness search admits amplitude amplification,
collision structure can support quantum walks, and sparse linear systems may
exhibit favorable quantum complexity under restrictive assumptions on access,
conditioning, state preparation, and output observability
\cite{shor1994algorithms,grover1996fast,brassard2000quantum,
ambainis2007quantum,harrow2009quantum}.
These results are often summarized by their headline speedups, yet the claim is
meaningful only relative to a precise computational model. An oracle-query
improvement may yield no end-to-end advantage when oracle construction, input
loading, reversible implementation, output recovery, or classical
post-processing dominates the total cost~\cite{aaronson2015read,tang2019quantum}.

Systematic analysis must preserve the original task while jointly checking its
classical bottleneck, access model, output contract, structural promises,
barriers, and defensible claim scope. These dependencies span classical
algorithms, quantum query models, and domain-specific formulations, making broad
and consistent coverage difficult.

Recent AI systems show that generation can usefully expand scientific and
algorithmic search when coupled to grounded feedback. Reinforcement-learning
systems have discovered matrix-multiplication procedures and low-level sorting
algorithms, while language-model-based systems have produced executable
mathematical constructions and formal geometry proofs
\cite{fawzi2022discovering,mankowitz2023faster,
romera2024mathematical,trinh2024solving}.
Scientific agents increasingly organize hypothesis formation, literature use,
experimentation, and critique into structured workflows
\cite{desai2025autoscilab,brahmavar2024generating,yang2025moose}.
In these settings, candidate evaluation is grounded in execution, formal
deduction, experiments, or machine-testable constraints that restrict
open-ended generation\cite{zhong2026agentodrl}.

Quantum-speedup analysis lacks an equally simple acceptance test.
Representative AI-for-quantum work addresses error correction, hardware
topology, laboratory control, and quantum programming
\cite{choukroun2024deep,li2025ai,cao2024agents,fu2025qagent}.
Executable benchmarks for quantum code generation further show that plausible
programs may remain semantically or algorithmically incorrect
\cite{guo2025quanbench,mikuriya2025qcoder}.
More broadly, evaluations of research agents reveal gaps between polished
research artifacts and performance under objective or replication-based
criteria
\cite{wijk2024re,starace2025paperbench,xiang2025scireplicate,zhang2026mlrc}.
For quantum algorithms, the central risk appears even earlier: a proposal may
be locally compatible with a known primitive while changing the task, assuming
unavailable coherent access, hiding the original computational burden inside
an oracle, or escalating a query-level improvement into an end-to-end claim~\cite{aaronson2015read}.

We therefore formulate AI-assisted quantum-speedup analysis as
\emph{bounded hypothesis generation}, rather than automated algorithm
discovery~\cite{zhou2024hypothesis,xiong2025toward,huang2025automated,fawzi2022discovering}. Generative models may propose structures, primitive matches,
complexity expressions, and candidate schemes, but they may not authorize or
strengthen their own conclusions~\cite{panickssery2024llm,xiong2025toward,kapoor2024ai}. Claims are instead bounded by immutable
public problem facts, a curated and source-linked primitive-and-barrier
registry, deterministic validation rules, and an explicit scope~\cite{dong2025xgrammar,lu2025learning,kapoor2024ai}. 
We call the
resulting support \emph{representation-relative}: it is evaluated against the
serialized public problem, registry, and stated scope, rather than an external
proof of algorithmic correctness. This creates an asymmetric trust
architecture: the model expands the candidate space, whereas deterministic
mechanisms restrict what each candidate may claim.

We instantiate this principle in \textbf{QuantumMind}. A deterministic
orchestrator executes nine typed actions for formalization, structure
analysis, primitive matching, barrier and prior-art assessment, scheme
generation, and bounded review. Each action has a strict output schema,
readable-context whitelist, and deterministic merge policy. A ten-check
validator assigns the authoritative verdict. Offline sidecars then compile a
\emph{Quantum Acceleration Evidence Graph} (QAEG) and apply a downward-only
research screen over output alignment, access upgrades, oracle risk, and
classical-baseline status. Neither sidecar can upgrade the decision.

On 582 paired tasks, QuantumMind reaches 53.1 mean ODS, compared with 35.8 for
the strongest of seven prompting and agentic controls. It wins 355 tasks, ties
98, and loses 129 against that baseline; passes the graph audit on 99.8\% of
tasks versus 43.6\%; and ranks first in each of seven task families. The
reviewer-only quality dimensions are much closer across systems, indicating
that the principal separation arises from production of validator-consistent,
auditable states rather than judge-facing fluency alone.

\noindent\textbf{Our contributions are threefold:}
\begin{itemize}
    \item We formalize quantum-speedup analysis as scoped,
    representation-relative hypothesis generation: a candidate is authorized
    only when its structure, access, output, promises, complexity certificate,
    barriers, and prior-art scope satisfy explicit registry obligations.

    \item We introduce QuantumMind, which separates typed agentic proposal
    generation from deterministic claim authorization, evidence-graph auditing,
    and downward-only research screening.

    \item We construct a paired open-discovery evaluation over 582 tasks and
    seven task-adapted controls. Under the frozen ODS rubric, QuantumMind ranks
    first in every task family and exceeds the strongest baseline by 17.3 points.
\end{itemize}

\section{Related Work}
\label{sec:related-work}

\subsection{AI for Scientific and Algorithmic Discovery}

AlphaTensor and AlphaDev search for improved algorithms using reinforcement
learning, whereas FunSearch couples language-model generation with executable
evaluation and AlphaGeometry combines neural proposals with symbolic deduction
\cite{fawzi2022discovering,mankowitz2023faster,
romera2024mathematical,trinh2024solving}.
Recent systems broaden this pattern to scientific workflows: AutoSciLab
supports interpretable laboratory discovery, logical-feedback methods impose
machine-testable chemical constraints, The AI Scientist automates substantial
parts of an ML research cycle, SciAgents uses ontological graph reasoning for
materials hypotheses, and MOOSE-Chem studies hypothesis rediscovery in
chemistry
\cite{desai2025autoscilab,brahmavar2024generating,
lu2024ai,ghafarollahi2025sciagents, yang2025moose}.
QuantumMind shares the goal of expanding a search space, but it does not assume
that one executable score establishes scientific validity. It instead records
both represented support and unresolved obligations.

\subsection{Agentic Workflows and Research Evaluation}

ReAct interleaves reasoning and tool use, while AutoGen and MetaGPT coordinate
multiple model calls through conversational or workflow structure
\cite{yao2022react,wu2023autogen,hong2024metagpt,li2026agentswift,zhao2026a2flow}.
Domain systems such as VerilogCoder and retrieval-enhanced drug-discovery
agents add execution or external evidence, and recent failure-localization work
argues for identifying which stage introduced an error
\cite{ho2025verilogcoder,lee2026rag,geng2026failure}.
Complementary benchmarks evaluate research agents against objective tasks,
replication rubrics, or executable implementations
\cite{wijk2024re,starace2025paperbench,xiang2025scireplicate}.
QuantumMind differs from unrestricted multi-agent dialogue: it uses a single
generic agent instantiated by a fixed table of typed actions, and agent
agreement never determines the authoritative verdict.

\subsection{AI for Quantum Computing and Auditable Claim Control}

AI-for-quantum research has addressed error correction, processor topology,
and autonomous laboratory operation
\cite{choukroun2024deep,li2025ai,cao2024agents}.
A growing adjacent literature studies LLM-based quantum programming and its
executable evaluation, including autonomous OpenQASM generation and benchmarks
for functional and semantic correctness
\cite{fu2025qagent,guo2025quanbench,mikuriya2025qcoder}.
These works generally assume that the target quantum program or task has
already been specified. QuantumMind instead operates upstream: it asks whether
a represented classical task satisfies the prerequisites of a registered
quantum primitive and what claim scope is supportable.

Knowledge graphs provide typed representations of entities, relations,
provenance, and evidence \cite{hogan2021knowledge}, and scientific agents have
used graph structure to retrieve cross-domain concepts and generate hypotheses
\cite{ghafarollahi2025sciagents,moreau2011open}.
QAEG has a narrower role. It is compiled deterministically after the ordinary
workflow, introduces no new evidence, and does not plan or rank with a learned
graph model. QuantumMind consequently separates three non-interchangeable
outputs: the authoritative validator verdict, graph-integrity status, and a
non-authoritative research disposition. A negative claim may have a valid
graph, and a representation-relative positive claim may still be demoted for
unresolved output, access, oracle, or baseline assumptions.

\section{Method}
\label{sec:method}

\begin{figure*}[t]
    \centering
    \includegraphics[width=\textwidth,keepaspectratio]{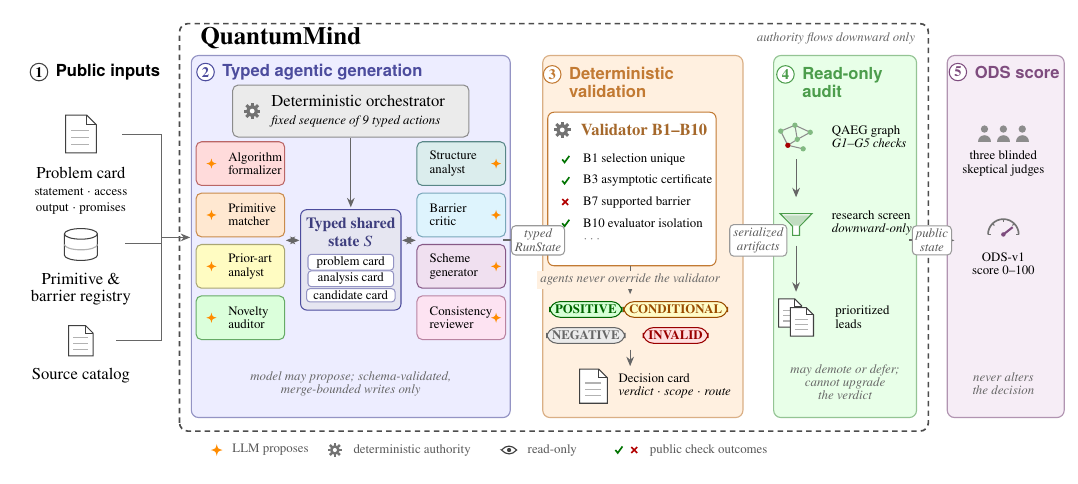}
    \caption{QuantumMind separates typed proposal construction from
    deterministic authorization. QAEG exposes the support behind the decision,
    and the research screen may only demote or defer a completed run. ODS-v1 is
    used only for evaluation.}
    \label{fig:quantummind-workflow}
\end{figure*}

Given a represented classical task, the system must make two related but
non-identical decisions: which registered quantum primitive is a plausible
match, and how strong a claim is justified by the stated access model, output,
promises, and costs. Language models are useful for proposing mappings and
schemes, but the same generative process should not authorize assumptions that
it introduced itself. QuantumMind therefore separates a typed \emph{proposal
plane} from a deterministic \emph{authorization plane}
(Fig.~\ref{fig:quantummind-workflow}).

Two further problems remain after authorization. First, a verdict alone does
not show how task facts, registry requirements, barriers, and complexity claims
support one another; QAEG exposes those dependencies. Second, an internally
consistent query-level claim may still be generic or impractical; a
research screen may therefore demote or defer it, but never strengthen it.
ODS is applied only afterward for evaluation. In short, agents may propose a
claim, while deterministic mechanisms control what the system is allowed to
retain.

\subsection{Typed Proposal Plane}

Specialized roles help decompose formalization, classical analysis, quantum
matching, and critique, but they must remain anchored to the same task. We
therefore use one shared typed state rather than free-form inter-agent dialogue.

Let \(x\) be the \emph{public task specification}: the statement, input and
access models, output contract, promises, size parameters, and unresolved
ambiguities.

The statement and model fields are copied once and never rewritten; the
formalizer may only append ambiguities. This prevents later agents from making
a proposal feasible by silently changing the problem being analyzed.

The workflow contains nine fixed actions. Let \(S_0\) be the initial state and
\(S_i\) the state after action \(i\in\{1,\ldots,9\}\):
\begin{equation}
\begin{aligned}
 S_i&=(x,A_i,C_i,M_i), \qquad i=0,\ldots,9,\\[-1mm]
 \textsc{None}&\prec\textsc{Query}\prec\textsc{Gate}
 \prec\textsc{EndToEnd}.
\end{aligned}
\label{eq:state-model}
\end{equation}
Here \(A_i\) is the analysis record, \(C_i\) the candidate record, and \(M_i\)
the append-only action history. The scope order is used later to prevent a
query-level primitive from being presented as a gate-level or end-to-end result.

Each action \(i\) has a prompt \(P_i\), output schema \(\mathcal Y_i\), readable
state view, deterministic update \(U_i\), and writable field set \(W_i\):
\begin{equation}
\begin{aligned}
 y_i&=\operatorname{Val}_{\mathcal Y_i}\!\left[
 L_\theta\!\left(P_i,\operatorname{View}_i(S_{i-1},\mathcal R)\right)
 \right],\\
 S_i&=U_i(S_{i-1},y_i),
 \qquad (S_i)_{\overline{W_i}}=(S_{i-1})_{\overline{W_i}}.
\end{aligned}
\label{eq:typed-transition}
\end{equation}
Schema violations and undeclared fields are rejected
\cite{dong2025xgrammar,lu2025learning}. Deterministic normalization may weaken
unsupported matches or literature claims but cannot strengthen them. The nine
actions proceed from formalization and structure analysis through primitive,
barrier, and prior-art assessment to scheme generation, critique, novelty, and
consistency review. Each call also records input, prompt, and output digests
with provider and parse metadata.

\subsection{Registry-Grounded Authorization}

Structural resemblance is insufficient for a quantum-speedup claim. A
search-like task may lack coherent oracle access, request an incompatible
output, or omit promises on which the complexity bound depends. The registry
makes these assumptions explicit rather than leaving them to persuasive prose.
For primitive \(p\), it records required structures \(Z_p\), allowed access
models \(A_p\), supported outputs \(O_p\), required promises \(\Pi_p\), maximum
scope \(s_p\), speedup class \(\kappa_p\), complexity statements, barriers, and
sources. It covers standard search, estimation, counting, walk, backtracking,
minimum-finding, period-finding, and linear-system pathways, together with
lower-bound diagnostics
\cite{grover1996fast,brassard2000quantum,brassard1998quantum,
ambainis2007quantum,durr1996quantum,montanaro2015quantum,
magniez2007search,shor1994algorithms,harrow2009quantum,
hoyer2002quantum,van1998quantum,beals2001quantum}.

Writing \(a(x)\), \(o(x)\), and \(\Pi(x)\) for the represented task model, and
\(Z(S)\) for the structures recorded in the analysis, the three central
compatibility predicates are
\begin{equation}
\begin{aligned}
 \operatorname{Sel}(p)
 &= [\kappa_p\neq\textsc{None}]\wedge[s_p\neq\textsc{None}],\\
 \operatorname{Mod}(p,x)
 &= [a(x)\in A_p]\wedge[o(x)\in O_p]
    \wedge[\Pi_p\subseteq\Pi(x)],\\
 \operatorname{Str}(p,S)&=[Z_p\subseteq Z(S)].
\end{aligned}
\label{eq:registry-conditions}
\end{equation}
Let \(m_p\) denote the agent's match label. A proposed plausible match that
fails \(\operatorname{Mod}\) is downgraded to \textsc{WeakAnalogy}; a
diagnostic registry entry becomes \textsc{NotSupported}. The generated
selection is retained only if
\begin{equation}
 p^\star=
 \begin{cases}
 p_{\rm gen},&
 \substack{\operatorname{Sel}(p_{\rm gen})\wedge
 \operatorname{Mod}(p_{\rm gen},x)\\
 \wedge[m_{p_{\rm gen}}=\textsc{Plausible}]},\\
 \bot,&\text{otherwise}.
 \end{cases}
\label{eq:path-normalization}
\end{equation}
Structure is deliberately checked again by B2, so an agent cannot certify its
own structural analysis. Constant-factor pathways cannot pass B3, and a
normalized no-candidate record clears scheme and quantum-complexity fields.

Barriers capture unresolved costs such as oracle construction, state
preparation, readout, and output recovery. Public access, output, or promise
facts may deterministically discharge a catalog barrier; otherwise the strongest
represented assessment is retained, with
\textsc{Supported} stronger than \textsc{Unknown}. A supported barrier blocks
only the scopes listed in the catalog, while an unresolved blocker makes the
result conditional rather than positive.

The authorization plane evaluates exactly ten public checks
(Table~\ref{tab:b-checks}).

\begin{table*}[t]
\centering
\small
\begin{tabular}{@{}p{1.05cm}p{1.75cm}p{12.6cm}@{}}
\toprule
Rules & Dimension & Deterministic requirement \\
\midrule
B1 & Selection & Exactly one selected \textsc{Plausible} registry match, or a clean no-candidate record. \\
B2 & Structure & The selected primitive exists and every required structure is represented. \\
B3 & Certificate & The registry class is \textsc{Asymptotic}; the scheme, the analysis and candidate classical baselines, and scope-relevant quantum complexity are present. \\
B4--B6 & Task model & Access and output are allowed, and every required promise is represented. \\
B7 & Barriers & No supported barrier blocks the claimed scope; an unresolved blocker yields \textsc{Unknown}. \\
B8 & Scope & The claim does not exceed the registry scope; a no-candidate record uses \textsc{None}. \\
B9 & Novelty & Global novelty is forbidden for known/direct prior art and unresolved when prior art is unknown. \\
B10 & Isolation & No evaluator-only field occurs in the public run state. \\
\bottomrule
\end{tabular}
\caption{B1--B10 are the only checks that assign the authoritative verdict.}
\label{tab:b-checks}
\end{table*}

Let \(b_i\) be the outcome of Bi. Define
\begin{equation}
\begin{aligned}
 F&=\bigvee_{i\in\{1,2,4,5,8,9,10\}}[b_i=\textsc{Fail}],\\
 N&=\operatorname{NoCand}(S)\vee[b_3=\textsc{Fail}]
    \vee[b_7=\textsc{Fail}],\\
 U&=\bigvee_{i\in\{3,6,7,9\}}[b_i=\textsc{Unknown}],\\[-1mm]
 V(S)&=
 \begin{cases}
  \textsc{Invalid},&C(S)=\emptyset\ \vee\ F,\\
  \textsc{Negative},&N,\\
  \textsc{Conditional},&U,\\
  \textsc{Positive},&\text{otherwise}.
 \end{cases}
\end{aligned}
\label{eq:verdict}
\end{equation}
Here \(\operatorname{NoCand}(S)\) denotes a normalized no-candidate record.
The first applicable branch is authoritative. Only a \textsc{Positive} result
retains the candidate scope, capped by \(s_p\); all other verdicts have scope
\textsc{None}. The route is then a fixed mapping to rerun, stop, review with
warnings, or expert review, and cannot upgrade the verdict.

\noindent\textbf{Registry-relative guarantee.}
Under correct registry labels, immutable public task fields, declared action
ownership, and evaluator isolation, \(V(S)=\textsc{Positive}\) implies one
selected plausible primitive whose structures, access, output, and promises are
represented; an asymptotic registry class and scope-relevant complexity
certificate; no supported or unresolved blocker at the claimed scope; no scope
escalation; and no forbidden novelty or evaluator field. This is a consistency
guarantee for the serialized claim, not a proof that the generated quantum
scheme is mathematically correct or end-to-end advantageous. The guarantee
follows directly from B1--B10 and the precedence in
Eq.~\eqref{eq:verdict}.

\subsection{QAEG: Deterministic Evidence-Graph Audit}

B1--B10 provide local outcomes, but their evidence is distributed across the
saved state. QAEG makes that dependency structure inspectable without assigning
a second scientific verdict. It first rejects a stale decision and then
compiles
\begin{equation}
\begin{aligned}
 \mathsf{Dec}_r&=\mathsf{Dec}(S_r),\\
 G_r&=\operatorname{Compile}(S_r,\mathsf{Dec}_r,\mathcal R)\\
    &=(\mathcal V_r,\mathcal E_r).
\end{aligned}
\label{eq:qaeg-short}
\end{equation}
Nodes represent task facts, structures, matches, registry entries, obligations,
barriers, schemes, complexities, novelty, claims, checks, and decisions; typed
edges encode support, discharge, scope bounds, and provenance. QAEG is a
deterministic projection, not a retrieval graph or learned judge
\cite{hogan2021knowledge,ghafarollahi2025sciagents}.

For the unique claim node \(v_c\), the displayed support subgraph is the least
fixed point reached by following only admissible incoming and outgoing evidence
edges:
\begin{equation}
 \mathcal C_r=
 \operatorname*{lfp}_{X\supseteq\{v_c\}}
 \left(X\cup\operatorname{Pred}_{\Lambda^-}(X)
          \cup\operatorname{Succ}_{\Lambda^+}(X)\right).
\label{eq:support-closure}
\end{equation}
Here \(\Lambda^-\) and \(\Lambda^+\) are the fixed edge-type sets permitted by
the verifier. This is a bounded dependency closure, not a global
minimum-cardinality subgraph. Field paths and action ownership localize missing
or contradictory support.

The five hard graph checks cover the claim-support path (G1), all expected
registry obligations (G2), barriers (G3), graph/state/decision integrity (G4),
and novelty scope (G5); G6 is diagnostic only. For outcomes
\(g_1,\ldots,g_5\), let
\(F_G=\bigvee_i[g_i=\textsc{Fail}]\) and
\(U_G=\bigvee_i[g_i=\textsc{Unknown}]\). Then
\begin{equation}
\begin{aligned}
 \Gamma(G_r)&=
 \begin{cases}
  \textsc{Fail},&F_G,\\
  \textsc{Warn},&\neg F_G\wedge U_G,\\
  \textsc{Pass},&\text{otherwise},
 \end{cases}\\
 A_G(r)&=[V(S_r)=\textsc{Positive}]\\[-1mm]
 &\quad\wedge[\Gamma(G_r)=\textsc{Pass}]
 \wedge\bigwedge_{i=1}^{5}[g_i=\textsc{Pass}].
\end{aligned}
\label{eq:graph-acceptance}
\end{equation}
The graph report \(R_r\) remains non-authoritative: a correctly represented
negative run may be graph-\textsc{Pass} while \(A_G(r)=0\).

\subsection{Downward-Only Research Screen}

Registry consistency is narrower than research usefulness. A valid query-level
claim may still be generic, depend on a black-box oracle, fail to recover the
original output, or use an uncertain classical baseline. The screen therefore
classifies output alignment, access upgrades, oracle construction, baseline
status, and candidate universe after QAEG. These labels remain
representation-relative: a described oracle is not a verified circuit, and
``best visible'' is not a proof of classical optimality. Generic wrappers are
demoted when they lack problem-specific structure or hide oracle cost
\cite{aaronson2015read,tang2019quantum}.

The first matching trigger is used in the following order: invalid state;
source mismatch; output mismatch or circular oracle; generic wrapper;
benchmark-only flag; unresolved output/access/oracle evidence or
\(A_G(r)=0\); unverified baseline; and default expert review. Let \(T_j\) be
these predicates, let \(T_8\equiv1\), and let
\((\ell_1,\ldots,\ell_8)\) denote the corresponding dispositions from
\textsc{InvalidState} through \textsc{KeepForExpertReview}. Then
\begin{equation}
 j^\star=\min\{j:T_j(S_r,G_r,R_r,\mathcal R)=1\},
 \qquad q_r=\ell_{j^\star}.
\label{eq:disposition-list}
\end{equation}
For the frozen artifacts \(z_r=(S_r,\mathsf{Dec}_r,G_r,R_r)\), the read-only
property is
\begin{equation}
\begin{aligned}
 \operatorname{Screen}(z_r)&=(z_r,q_r),\\
 q_r=\ell_8&\Rightarrow[\Gamma(G_r)=\textsc{Pass}]\\[-1mm]
 &\quad\wedge[A_G(r)=1].
\end{aligned}
\label{eq:authority-preservation}
\end{equation}
The screen may therefore demote or defer a run, but it cannot change its
verdict, scope, route, graph status, or acceptance indicator.
\section{Experiments}
\label{sec:experiments}

\subsection{Protocol}

We evaluate all systems on the same 582 open-discovery tasks. These tasks are
the uniquely matched subset of 628 completed runs; 46 runs belonging to 23
duplicate-card collision groups are excluded rather than paired by an
arbitrary ordering. Each method receives the same public task and registry
views, emits the same typed artifacts, and is passed through the same
B1--B10 validator, QAEG audit, downward-only screen, and ODS-v1 evaluator.
The resulting grid contains 4,656 system--task records, all of which complete
scoring successfully.

The controls span the main interaction patterns used by recent agentic
systems: direct and staged prompting, self-consistency, interleaved
reasoning--acting, proposer--critic dialogue, bounded ideation and review, and
graph-grounded multi-agent reasoning
\cite{wang2022self,yao2022react,wu2023autogen,
hong2024metagpt,lu2024ai,ghafarollahi2025sciagents,wei2022chain,du2023improving}. They are
task-adapted controls under the common artifact contract rather than full
reproductions of the original domain-specific systems. This paired design
follows the broader agent-evaluation principle of separating a plausible
generated artifact from success under an objective or executable assessment
\cite{ho2025verilogcoder,starace2025paperbench,
xiang2025scireplicate,kapoor2024ai,siegel2024core}.

\begin{table*}[t]
\centering
\small
\begingroup
\renewcommand{\arraystretch}{0.96}
\begin{tabular*}{\textwidth}{@{\extracolsep{\fill}}lrrrrrrrr@{}}
\toprule
\multicolumn{9}{c}{\textbf{Panel A: Overall quality and evidence control}}\\
\cmidrule(lr){1-9}
System & ODS$\uparrow$ & \(\Delta_{\rm QM}\) & \(T/E/R\) &
Audit$\uparrow$ & Claim & Invalid$\downarrow$ & Strong$\uparrow$ &
QM W/T/L\\
\midrule
\textbf{QuantumMind}
& \textbf{53.1} & -- & 2.03/2.14/2.05
& \textbf{99.8} & \textbf{66.0} & \textbf{0.2} & \textbf{0.3} & --\\
Direct
& 23.9 & 29.2 & 2.05/2.10/2.03
& 10.0 & 0.2 & 88.8 & 0.0 & 525/24/33\\
Staged
& 35.4 & 17.7 & 2.02/2.07/2.01
& 42.4 & 0.0 & 57.6 & 0.0 & 357/89/136\\
Self-consistency
& \underline{35.8} & \underline{17.3} & 2.01/2.07/2.01
& \underline{43.6} & 0.0 & \underline{56.4} & 0.0 & 355/98/129\\
ReAct-style
& 35.3 & 17.8 & 2.05/2.19/2.04
& 40.7 & 0.0 & 59.3 & 0.0 & 364/65/153\\
Proposer--critic
& 20.6 & 32.5 & 2.07/2.27/2.07
& 0.3 & 0.2 & 99.7 & 0.0 & 580/1/1\\
Ideate--review
& 22.6 & 30.5 & 2.09/2.22/2.09
& 5.3 & 0.2 & 93.5 & 0.0 & 548/13/21\\
SciAgents-style
& 21.0 & 32.1 & 2.04/2.18/2.03
& 2.4 & 0.0 & 97.6 & 0.0 & 568/8/6\\

\midrule
\multicolumn{9}{c}{\textbf{Panel B: Mean ODS by task family}}\\
\cmidrule(lr){1-9}
System & Search$_{326}$ & Estim.$_{30}$ & Graph$_{198}$ & Count.$_{12}$ &
Backtr.$_{8}$ & Q-walk$_4$ & Min.$_4$ & Mean\\
\midrule
\textbf{QuantumMind}
& \textbf{50.6} & \textbf{49.6} & \textbf{56.4} & \textbf{62.6}
& \textbf{60.7} & \textbf{65.5} & \textbf{60.0} & \textbf{53.1}\\
Direct
& 22.9 & 23.8 & 25.0 & 25.6 & 24.5 & 28.7 & \underline{39.7} & 23.9\\
Staged
& 34.4 & 32.3 & 38.0 & \underline{28.9} & 33.9 & 28.7 & 37.9 & 35.4\\
Self-consistency
& 33.3 & \underline{37.1} & \underline{40.3} & 28.7
& \underline{37.9} & \underline{37.9} & 28.7 & \underline{35.8}\\
ReAct-style
& \underline{36.3} & \underline{37.1} & 34.7 & 22.9 & 24.0 & 32.3 & 29.5 & 35.3\\
Proposer--critic
& 20.6 & 20.4 & 20.6 & 20.3 & 19.4 & 23.6 & 20.8 & 20.6\\
Ideate--review
& 22.6 & 22.4 & 22.7 & 23.2 & 20.5 & 30.8 & 21.2 & 22.6\\
SciAgents-style
& 20.8 & 19.8 & 21.7 & 19.4 & 19.4 & 20.8 & 20.8 & 21.0\\
\bottomrule
\end{tabular*}
\endgroup
\caption{Paired results on 582 identical tasks. Panel A reports overall ODS
and evidence-control outcomes. QM abbreviates QuantumMind;
\(\Delta_{\rm QM}\) is its mean ODS advantage over the row system, and QM
W/T/L counts its task-level wins, ties, and losses against the row system.
\(T/E/R\) are the reviewer dimensions. Claim is the accepted-claim rate,
Audit is the graph-PASS rate, Invalid is the \textsc{InvalidState} rate, and
Strong is the ODS-$\geq70$ rate. Panel B uses the same method-as-row organization
and reports mean ODS by task family; header subscripts give task counts, and the
final column repeats the overall mean. Bold denotes the best result and
underlining the strongest baseline at displayed precision.}

\label{tab:main-agentic-comparison}
\end{table*}
\subsection{Open-Discovery Evaluation}

For run \(r\), reviewer \(j\in\{1,2,3\}\) assigns
\(y_{rjk}\in\{0,1,2,3,4\}\) for
\(k\in\{T,E,R\}\), corresponding to technical validity, epistemic
auditability, and research utility. We aggregate the three reviews using their
median and median absolute deviation \cite{rousseeuw1993alternatives}. With
\(\epsilon=0.02\),
\begin{align}
x_{rjk}
&=\operatorname{clip}(y_{rjk}/4,\epsilon,1-\epsilon),
\nonumber\\
m_{rk}
&=\operatorname{med}_{j}x_{rjk},
\qquad
d_{rk}=\operatorname{med}_{j}|x_{rjk}-m_{rk}|,
\nonumber\\
\widetilde{x}_{rk}
&=\operatorname{clip}\!\left(
m_{rk}-0.7413d_{rk},\epsilon,1-\epsilon\right),
\nonumber\\
L_r
&=\widetilde{x}_{rT}^{0.4}
  \widetilde{x}_{rE}^{0.4}
  \widetilde{x}_{rR}^{0.2}.
\label{eq:ods-semantic-exp}
\end{align}

Let \(q_r\in\{\ell_1,\ldots,\ell_8\}\) be the ordered disposition in
Eq.~\eqref{eq:disposition-list}, let \(a_r=A_G(r)\), and let
\(g_r=\mathbf{1}[\Gamma(G_r)=\textsc{Pass}]\). The frozen disposition
anchors are
\begin{equation}
\bigl(\pi(\ell_1),\ldots,\pi(\ell_8)\bigr)
=(.04,.10,.12,.36,.45,.70,.84,.95).
\label{eq:ods-prior-exp}
\end{equation}

The deterministic prior and semantic fusion are
\begin{align}
D_r
&=\sigma\!\left(
  \operatorname{logit}\pi(q_r)+0.5a_r-1.5(1-g_r)\right),
\nonumber\\
P_r
&=\sigma\!\left(
  0.7\operatorname{logit}L_r+0.3\operatorname{logit}D_r\right).
\label{eq:ods-fusion-exp}
\end{align}

Define
\begin{align}
f_r
&=\mathbf{1}\!\left[
\Gamma(G_r)=\textsc{Fail}\vee
q_r\in\{\ell_1,\ell_2,\ell_3\}\right],
\nonumber\\
u_r&=\mathbf{1}[q_r=\ell_4],
\qquad
b_r=\mathbf{1}[q_r=\ell_5],
\nonumber\\
C_r
&=\min\{1,0.25^{f_r},0.58^{u_r},
          0.65^{b_r},0.68^{1-a_r}\}.
\label{eq:ods-cap-exp}
\end{align}

With
\(\operatorname{sp}(z)=\max(z,0)+\log(1+e^{-|z|})\),
\begin{equation}
\operatorname{ODS}(r)=
100\operatorname{clip}\!\left[
P_r-\frac{\operatorname{sp}(40(P_r-C_r))}{40},0,1\right].
\label{eq:ods-final-exp}
\end{equation}

For task set \(\mathcal T\), we report
\begin{align}
\operatorname{MeanODS}(s)
&=\frac{1}{|\mathcal T|}
  \sum_{t\in\mathcal T}\operatorname{ODS}(s,t),
\nonumber\\
\Delta_{\mathrm{QM},b}
&=\frac{1}{|\mathcal T|}
  \sum_{t\in\mathcal T}
  \left[\operatorname{ODS}(\mathrm{QM},t)-
        \operatorname{ODS}(b,t)\right],
\nonumber\\
\operatorname{Lead}_{\tau}(s)
&=\frac{1}{|\mathcal T|}
  \sum_{t\in\mathcal T}
  \mathbf{1}[\operatorname{ODS}(s,t)\geq\tau],
\;
\tau\in\{70,85\}.
\label{eq:ods-aggregate-exp}
\end{align}

The geometric semantic term prevents compensation across dimensions, whereas
the prior and cap prevent fluent text from overriding an invalid graph state,
an unaccepted claim, or a generic-wrapper diagnosis. ODS is a
rubric-calibrated research-utility score, not a correctness or novelty
probability; the blinded, multi-reviewer design addresses known LLM-judge
biases but does not eliminate them
\cite{shi2025judging,wang2024large}.

\subsection{Results}

\paragraph{Overall comparison.}
QuantumMind obtains a mean ODS of 53.1, exceeding the strongest baseline,
self-consistency, by 17.3 points (48.2\% relative). The advantage is broad
rather than driven by a few tasks: against self-consistency, QuantumMind wins
355 tasks, ties 98, and loses 129; across all seven controls it wins between
355 and 580 of the 582 paired tasks. All controls complete scoring, so these
gaps are not caused by missing evaluations.

The reviewer-only dimensions are close: system means range from 2.01 to 2.27
on the 0--4 \(T/E/R\) scales, and several dialogue baselines slightly exceed
QuantumMind on individual sub-scores. The largest separation appears after
deterministic evidence control: QuantumMind achieves a 99.8\% graph-PASS rate
and 66.0\% claim acceptance, whereas the strongest baseline reaches 43.6\%
and 0.2\%, respectively. Its \textsc{InvalidState} rate is 0.2\%, compared
with 56.4--99.7\% for the controls. The empirical advantage is therefore
concentrated in producing artifacts that remain coherent under the shared
validator and graph audit.

Among the controls, staged prompting, self-consistency, and ReAct-style
interaction form the strongest group (35.3--35.8 ODS). Direct prompting
reaches 23.9, while proposer--critic, ideate--review, and SciAgents-style
dialogue reach 20.6--22.6. More interaction therefore does not by itself
improve performance under a strict scientific artifact contract: without
typed field ownership and validator-aligned state transitions, additional
critique often remains semantically plausible but structurally invalid.

\paragraph{Task-family behavior and an illustrative run.}
QuantumMind ranks first in every task family, while the strongest control varies
by structure (Table~\ref{tab:main-agentic-comparison}). Search and estimation
retain represented query-level claims but are uniformly demoted as generic
wrappers; graph/local tasks instead yield well-formed negative or reformulation
records. The four structured families score 60.0--65.5 and produce the only two
ODS-$\geq70$ leads, one in counting and one in marked-state walk search.

Figure~\ref{fig:illustrative-run} makes this separation concrete using an
original-cohort counting run, selected for compactness and interpretability
rather than maximum ODS. Its query-level proposal passes deterministic
authorization and graph auditing, yet the research screen still withholds the
strongest disposition. These outputs are intentionally non-equivalent: the
former certifies the represented claim, whereas the latter preserves unresolved
scientific obligations.

\begin{figure}[t]
\centering
\begingroup
\small
\fbox{\parbox{\dimexpr\columnwidth-2\fboxsep-2\fboxrule\relax}{%
\raggedright
{\centering
\textbf{Illustrative Run: Approximate Marked-Set Counting}\par
\emph{Original paired cohort; ODS 65.8}\par}
\smallskip\hrule\smallskip

\textbf{Input.}
Given an indexed set of $N$ candidates and, under the probe's assumptions,
coherent access to a Boolean predicate $f$, estimate
$M=\sum_{x=0}^{N-1}f(x)$, or the marked fraction $p=M/N$, to additive
precision. The requested output is a scalar estimate, not the marked objects.\par
\smallskip

\textbf{System proposal.}
The detected structure is marked-set cardinality estimation and the selected
primitive is \texttt{quantum\_counting}. The scheme prepares the uniform state,
applies the Grover counting iterate with phase-estimation-style counting, and
returns $\widehat M=N\widehat p$. For fraction error $\epsilon$ and failure
probability $\delta$, the saved black-box query comparison is\par
\smallskip

\begin{tabular}{@{}ll@{}}
Classical: & $O\!\left(\min\{N,\epsilon^{-2}\log(1/\delta)\}\right)$,\\
Quantum:   & $O\!\left(\epsilon^{-1}\log(1/\delta)\right)$.
\end{tabular}\par

For raw-count error $\Delta$, the proposal sets $\epsilon=\Delta/N$. The
claimed scope is \textsc{Query}.\par
\smallskip\hrule\smallskip

\textbf{Deterministic authorization.}
All ten B checks pass. The authoritative verdict is \textsc{Positive}, the
maximum supported scope is \textsc{Query}, and the route is
\textsc{Expert Review}. QAEG is \textsc{Pass} and accepts the represented
claim.\par
\smallskip

\textbf{Downward screen.}
The final disposition is nevertheless \textsc{Reformulate}: output sufficiency,
coherent-access justification, reversible oracle construction, and the best
classical baseline remain unresolved.\par
\smallskip

\textbf{Supported conclusion.}
The run supports a registry-relative query-complexity hypothesis for approximate
counting. It does not establish a gate-level or end-to-end speedup for the
parent combinatorial problem.
}}
\caption{Original-cohort run FDV1-00009, abridged from its saved artifacts.
Prose is condensed for readability; all displayed bounds, statuses, and
decision fields are preserved without strengthening the claim.}
\label{fig:illustrative-run}
\endgroup
\end{figure}

\paragraph{Takeaway.}
Under the frozen evaluator, QuantumMind's main advantage is a substantially higher rate of scientifically
well-formed states. The comparison therefore supports the intended role of
the architecture: generation expands candidate hypotheses, while typed state
transitions, deterministic validation, graph auditing, and downward-only
screening determine which parts of those hypotheses remain supportable.

\section{Limitations}
\label{sec:limitations}

QuantumMind evaluates represented quantum-acceleration hypotheses rather than
replacing formal proof, implementation-level resource analysis, or expert
review. Coverage depends on the public task representation and the primitives,
barriers, and sources encoded in the registry; extending these resources is a
straightforward route to broader domains. The present suite also emphasizes
query- and subroutine-level opportunities, while an end-to-end claim may
require additional analysis of loading, oracle construction, reversible
implementation, readout, and the strongest classical baseline\cite{aaronson2015read,tang2019quantum}.

The compared systems do not use identical inference budgets. QuantumMind
executes a fixed sequence of specialized prompts, whereas some controls use a
single call and others use a smaller number of sampled or interactive calls.
The results therefore evaluate complete system designs rather than isolating
orchestration from prompt count or token budget. Multi-call controls reduce
this gap, but a strictly call- or token-matched ablation remains useful future
work. Finally, ODS is a frozen measure of auditability and research utility;
it prioritizes records for follow-up and is not a probability of correctness
or novelty\cite{kapoor2024ai,yang2026bamas}.

\section{Conclusion}
\label{sec:conclusion}

We introduced QuantumMind, an agentic workflow for producing structured,
source-aware, and auditable hypotheses about quantum acceleration. It preserves
public task facts in a typed shared state, while deterministic B1--B10
validation, QAEG auditing, and a downward-only research screen control what may
be claimed. This separation keeps model, scope, barrier, and provenance
obligations explicit without treating fluent generation as proof of quantum
advantage.

Across 582 paired open-discovery tasks, QuantumMind ranks first in all seven
task families and reaches 53.1 mean ODS, 17.3 points above the strongest
control. It records 355 wins, 98 ties, and 129 losses, and passes the graph
audit on 99.8\% of tasks versus 43.6\% for that baseline. The main empirical
advantage is artifact-level consistency rather than more persuasive prose.

Future work will broaden the primitive and barrier registries, add
implementation-aware resource estimation and formal checking, and connect
high-scoring records to blinded expert review and executable prototypes.
Call- and token-matched ablations, cross-model replication, and independent
ODS calibration should better isolate the sources and transferability of these
gains. Oracle-construction analysis, gate-level resource models, and
literature-grounded prior-art search could expose hidden assumptions. The same
interface can incorporate new primitives, barriers, and evaluators without
allowing generative components to strengthen the authoritative verdict.

\bibliography{aaai2027}

\end{document}